\documentclass[lettersize,journal]{IEEEtran}
\usepackage{amsmath,amsfonts}
\usepackage{algorithmic}
\usepackage{array}
\usepackage[caption=false,font=normalsize,labelfont=sf,textfont=sf]{subfig}
\usepackage{textcomp}
\usepackage{stfloats}
\usepackage{url}
\usepackage{verbatim}
\usepackage{graphicx}

\usepackage{graphicx}
\usepackage{multirow}
\usepackage{bbding}
\usepackage{tikz}

\def\BibTeX{{\rm B\kern-.05em{\sc i\kern-.025em b}\kern-.08em
    T\kern-.1667em\lower.7ex\hbox{E}\kern-.125emX}}
\usepackage{balance}
\begin{document}
\title{Produce Once, Utilize Twice for Anomaly Detection}
\author{Shuyuan Wang, Qi Li, Huiyuan Luo, Chengkan Lv, Zhengtao Zhang
\thanks{Manuscript created November, 2023; 
This work has been submitted to the IEEE for possible publication. Copyright may be transferred without notice, after which this version may no longer be accessible. 

Shuyuan Wang, Qi Li, Huiyuan Luo, Chengkan Lv, and Zhengtao Zhang are with the Institute of Automation, Chinese Academy of Sciences, Beijing 100190, China (e-mail: wangshuyuan2020@ia.ac.cn, qi.li@ia.ac.cn, huiyuan.luo@ia.ac.cn; chengkan.lv@ia.ac.cn; zhengtao.zhang@ia.ac.cn).
}}

\markboth{}%
{}

\maketitle

\begin{abstract}
Visual anomaly detection aims at classifying and locating the regions that deviate from the normal appearance. Embedding-based methods and reconstruction-based methods are two main approaches for this task.
The embedding-based methods typically predict the anomaly by measuring the distances between the deep representations of the test samples and a limited number of nominal samples, which enables these methods to be efficient but struggle in providing a fine-grained pixel-level anomaly location.
The reconstruction-based methods rely on the pixel-level reconstruction errors to locate the anomaly, thereby the anomaly predictions are fine-grained.
However, there are repetitive feature extractions and usually extra modules to guarantee the quality of the reconstructed images, resulting in unsatisfactory detection efficiency.
In a nutshell, the prior methods are either not efficient or not precise enough for the industrial detection.
To deal with this problem, we derive POUTA (Produce Once Utilize Twice for Anomaly detection), 
which improves both the accuracy and efficiency by reusing the discriminant information potential in the reconstructive network.
We observe that the encoder and decoder representations of the reconstructive network are able to stand for the features of the original and reconstructed image respectively.
And the discrepancies between the symmetric reconstructive representations provides roughly accurate anomaly information.
To refine this information, a coarse-to-fine process is proposed in POUTA, which calibrates the semantics of each discriminative layer by the high-level representations and supervision loss.
Equipped with the above modules, POUTA is endowed with the ability to provide a more precise anomaly location than the prior arts.
Besides, the representation reusage also enables to exclude the feature extraction process in the discriminative network, which reduces the parameters and improves the efficiency. Extensive experiments show that, POUTA is superior or comparable to the prior methods with even less cost. 
Furthermore, POUTA also achieves better performance than the state-of-the-art few-shot anomaly detection methods without any special design, showing that POUTA has strong ability to learn representations inherent in the training data.

\end{abstract}

\begin{IEEEkeywords}
Anomaly detection, detection efficiency, representations reusing.
\end{IEEEkeywords}

\section{Introduction}
\label{sec:intro}

\begin{figure}[t]
\begin{center}
    \includegraphics[width=\linewidth]{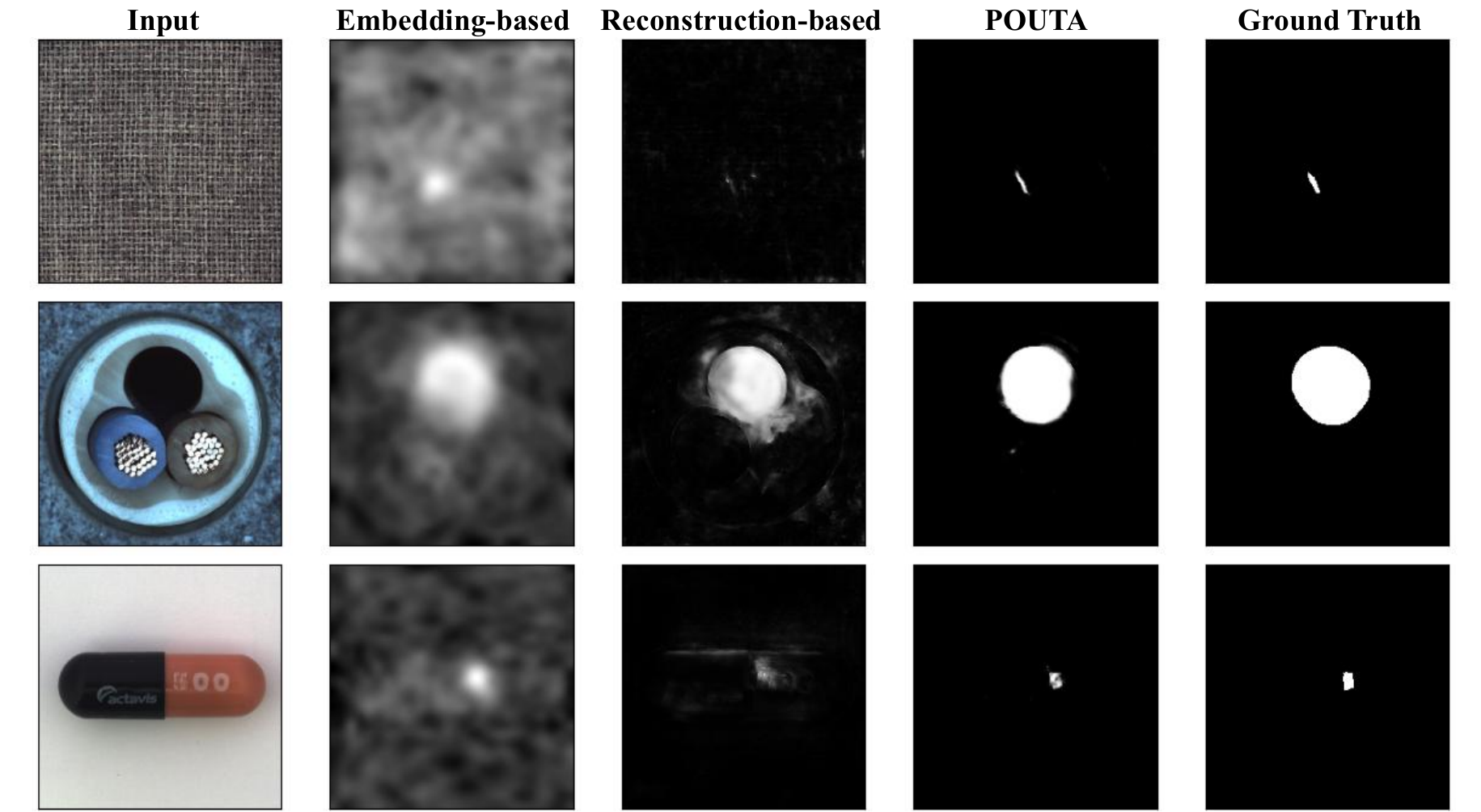}
\end{center}
   \caption{The qualitative examples of the reconstruction-based (DR{\AE}M \cite{DRAEM}), embedding-based (PatchCore \cite{PatchCore}) methods and the proposed POUTA.}
\label{fig:intro1}
\end{figure}

\IEEEPARstart{U}{nsupervised}
anomaly detection has drawn much attention in recent years~\cite{GOOD,reverse}, since it requires no defect samples for training \cite{NIGSF}, which benefits the scenario where the collection of defect samples is costly, e.g. industrial detection. 
There are two main approaches for anomaly detection, reconstruction-based methods and embedding-based methods. 
The reconstruction-based methods train a reconstructive network to build a normal version of the input. Generally, the deviation lies solely in the abnormal regions before and after reconstruction. By analyzing the reconstruction errors, the anomaly can be located \cite{GOOD, GANomaly, skipGANomaly}. 
As illustrated in Figure \ref{fig:intro1}, due to the analysis on pixel-level reconstruction errors, the reconstruction-based methods are endowed with the capability to locate the anomaly at the fine-grained pixel-level.
However, there are strict requirements for the reconstructive network. Specifically, abnormal regions should be repaired to normal, while the normal regions are supposed to remain unchanged. Otherwise, the reconstruction errors will not conform to the anomaly, resulting in failure in detection, as shown in the carpet of Figure \ref{fig:intro1}. 
Prior studies \cite{Memory1, AFEAN, DBPI} always design extra modules for the reconstructive network to meet the aforementioned requirements.
Due to the intricate structure, the reconstruction-based methods fall flat when comparing the inference time with the embedding-based methods. 
The embedding-based methods typically exploit the deep representations to do a feature matching between the test sample and the nominal samples.
Since the nominal samples for feature matching are usually limited in number \cite{PatchCore}, the embedding-based methods achieve tremendous progress on the detection efficiency. However, the deep representations are typically of lower resolution than the input, which makes the embedding-based methods to struggle in providing a fine-grained pixel-level anomaly location, as shown in Figure \ref{fig:intro1}.
In a nutshell, 
both of the above methods have some deficiencies, 
and it is an ambitious goal for the prior arts to provide a fine-grained pixel-level anomaly location efficiently. 

There are two approaches to overcome the aforementioned deficiencies. 
For the embedding-based methods, to obtain the fine-grained location, it is necessary to analyze the shallow representations to add details about the anomaly. However, modeling or feature matching the shallow representations is sophisticated. 
Since the shallow representations are not as homogeneous as the deep ones, it requires to record the shallow representations elaborately to obtain an accurate anomaly location, which is costly and may deprive the efficiency. 
Therefore, we turn to another solution, improving the accuracy and efficiency of the reconstruction-based methods. 

Recent reconstruction-based methods follow a paradigm, which involves a reconstructive network and a discriminative network~\cite{RSTD,DRAEM,Lv}. The discriminative network is employed to locate the anomaly by analyzing the discrepancies between the original and reconstructed images, demonstrating significantly better performance than simply measuring the absolute distance between these two images. 
Despite the impressive efforts, two critical issues still exist. Firstly, this paradigm yet relies on the reconstruction errors to identify anomaly. The requirements for the reconstructive network are strict as usual. While the quality of the reconstructed image is sometimes not so ideal, resulting in the unsatisfactory prediction, as shown in the carpet and capsule of Figure \ref{fig:intro1}. Even if the reconstructed image is of high quality, it is still a challenging problem to indicate anomaly completely when the color of abnormal region is close to the normal one.
The second lies in the excessive computing consumption caused by the repetitive feature extraction in the reconstructive and discriminative network. 
Both of them extracts the features of the input in their encoders.
It seems impossible to solve both of the above problems simultaneously within the conventional reconstruction-based paradigm, as achieving better performance necessitates enhancing the quality of reconstruction, which inevitably leads to increased computing consumption. This enlightens us to propose a new reconstruction-based paradigm.

\begin{figure}[t]
\begin{center}
    \includegraphics[width=\linewidth]{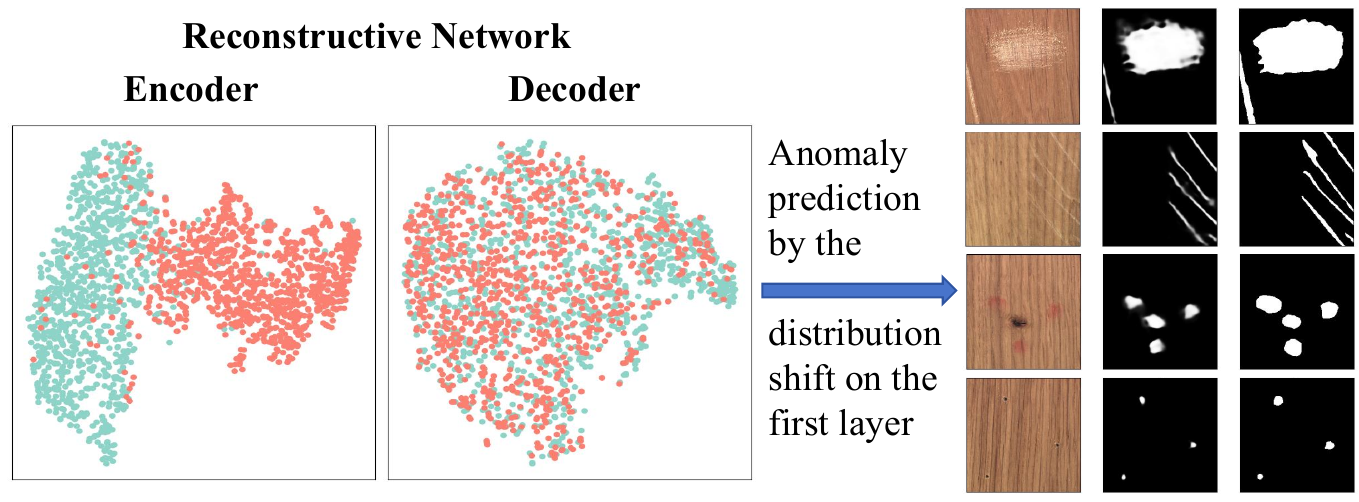}
\end{center}
   \caption{In POUTA, the representation distributions of the last reconstructive encoder layer and the first reconstructive decoder layer are visualized by t-SNE \cite{tSNE}. Each column on the right represents the input, the anomaly predicted only by the differences in representation of the aforementioned two layers, and the ground truth, respectively.}
\label{fig:intro2}
\end{figure}

We argue that, in the reconstructive network, the encoder and decoder representations are able to stand for the hierarchical representations of the input and reconstructed images. It is more suitable 
for the discriminative network to locate the anomaly by analyzing the discrepancies between the symmetric representations in the encoder and decoder directly.
To verify our assumption, the encoder and decoder representations are visualized in Figure~\ref{fig:intro2}. Figure~\ref{fig:intro2} indicates that, in the encoder, most of the abnormal representations are apart from the normal ones. While in the decoder, since the abnormal regions are repaired to normal, their representations are shifted to the distribution of the normal data. 
The above phenomena are consistent with our assumption. Since the normal regions and the abnormal regions in the input have different appearances, they naturally belong to different distributions. While the abnormal regions are repaired to normal in the reconstructed image, their representations will shift to the normal distribution in the decoder. 
Therefore, it is feasible to identify the anomaly by the representation discrepancies between encoder and decoder. As shown in Figure ~\ref{fig:intro2}, by analyzing the disparities between the last reconstructive encoder layer and the first reconstructive decoder layer, the anomaly can be located roughly accurately. 

\begin{figure*}[h]
\begin{center}
    \includegraphics[width=0.8\textwidth]{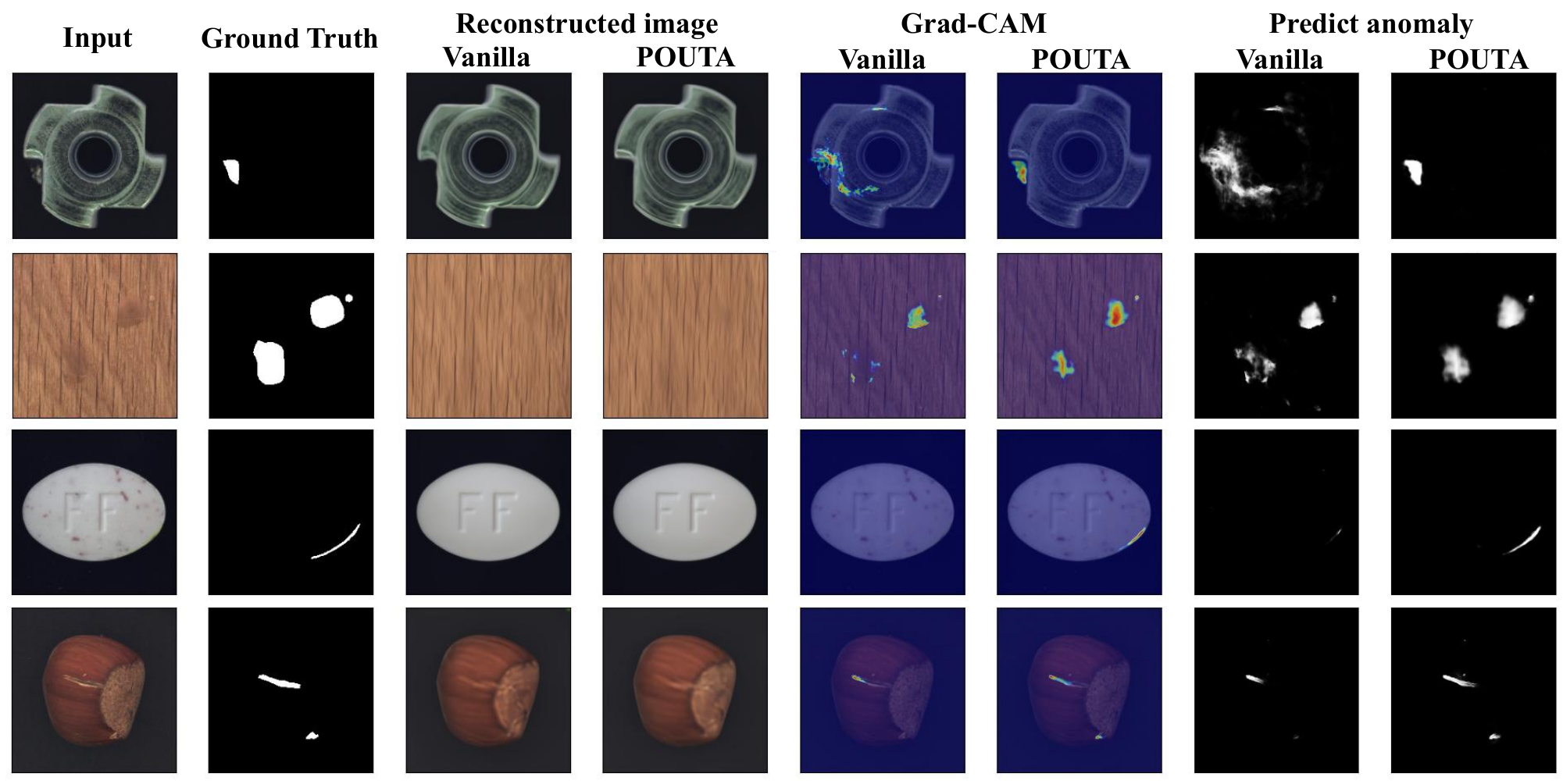}
\end{center}
   \caption{The qualitative examples and visualizations (provided by Grad-CAM \cite{gradcam}) of the vanilla reconstruction-based (DR{\AE}M \cite{DRAEM}-like) and POUTA.}
\label{fig:intro3}
\end{figure*}

Based on the above conclusion, we derive a neat and effective approach, POUTA (Produce Once Utilize Twice for Anomaly detection), as an effective remedy for the reconstruction-based methods on both accuracy and efficiency, fully reusing the discriminant information potential in the reconstructive network.
POUTA is established with a view to the availability of the multi-scale reconstructive representations, aggregating hierarchical symmetric representations to identify anomaly through a coarse-to-fine discriminative network. 
Specifically, in the discriminative network, firstly, there is a MSS (Multi-scale Semantic Supervision) module to ensure clear differentiation for the abnormal regions between encoder and decoder representations. Secondly, a differential module named FCM (Feature Contrast Module) is designed to analyze the distribution shift of the same representation between the encoder and decoder in the feature space. 
The representations of the normal regions remain in the same distribution, while those of the abnormal ones shift, which can serve as a signal for detection. 
In addition, HSG (High-level Semantic Guidance) module calibrates the hierarchical representations output by FCM progressively, resulting in rich and precise spatial details.

Due to reusing the reconstructive representations, POUTA suffices to identify anomaly more precisely than the vanilla reconstruction-based method with even lower cost.
We take Grad-CAM \cite{gradcam} to visualize the semantic differences captured by the vanilla method and POUTA in Figure \ref{fig:intro3}. There is no significant discrepancy between the reconstructed images from both methods, while POUTA obtains more complete semantics of anomaly regions through the representations and therefore the predictions of POUTA are more precise. This indicates that the reconstructive representations are more informative than reconstruction errors.
It is reasonable since the information loss occurs when reconstructing an image from its representations. 
As a result of this loss, the pixel-level reconstruction errors might be incomplete, or even inaccurate, potentially leading to detection failures in certain cases, especially those of fine-grained anomalies.
As shown in Figure \ref{fig:intro3}, the colors of the abnormal regions in wood and pill are close to the normal ones, the vanilla method fails to locate anomalies without surprise, but POUTA still identifies anomaly precisely.
Besides, the discriminative network in POUTA directly reuses the reconstructive representations, which distinctively allows for the exclusion of feature extraction in the discriminative network, thereby improving efficiency. In summary, by directly reusing informative representations, POUTA relaxes the strict requirement for reconstructive network, and this new paradigm is able to achieve higher accuracy at lower cost.

Our main contributions can be listed as follows:
\begin{itemize}
    \item  Different from previous reconstruction-based methods, 
    the discriminative network in POUTA directly reuses the representations in the reconstructive network, which enables it to be more efficient.
    \item POUTA utilizes the reconstructive representations instead of the reconstructed image to locate the anomaly, which provides more accurate and complete information about the abnormal regions, resulting in a more precise anomaly location.
    \item Experimental results show the superiority of the proposed unsupervised method POUTA. It outperforms the state-of-the-art methods on MVTec AD and VisA dataset and achieves a comparable performance with the best-performing supervised method on DAGM dataset. 
    \item  POUTA also achieves superior performance on few-shot anomaly detection without any special design, which indicates that POUTA has strong ability to learn representations inherent in the nominal data.
\end{itemize}

\section{Related Work}
Anomaly detection is an unsupervised detection task, where only the normal data is involved amid training. The anomaly detection model is required to distinguish between samples being drawn from the training data distribution and those outside its support \cite{PatchCore}.
Given respective nominal samples and test samples, anomaly detection can be converted to calculating distances between the nominal deep representations and the test ones, or analyzing the differences before and after reconstruction, which corresponds to the embedding-based methods and reconstruction-based methods.

\noindent \textbf{Embedding-based methods.} The embedding-based methods typically extract the deep representations and then calculate the distance to $k$ 
nearest nominal neighbors or the probability in the distribution.
DeepSVDD~\cite{DeepSVDD} extracts the deep representations describing the entire image, which limits the model to obtaining only the existence of the anomaly but not to locating it. To mitigate this problem, PatchSVDD \cite{PatchSVDD} extracts the deep representations for the image patch, which provides the rough anomaly location. 
Similarly, PaDim \cite{Padim} generates patch embeddings from the pretrained network to estimate the Gaussian distribution moments (mean and covariance) for patch-level Mahalanobis distance.
Other distribution learning methods can also be utilized, such as the  normalizing flow (NF) \cite{NF}.
DifferNet \cite{DifferNet} takes the NF to measure the anomaly. To obtain the more informative representations, CSflow \cite{CSflow} builds a cross-scale NF, which enables the model to predict the anomaly more precisely while increase the model scale at the same time. 
Some works turn the anomaly detection task into a matter of distances to $k$ nearest neighbors.
SPADE records the nominal representations extracted from a pretrained network in a memory bank, and then calculates the average distances between the test data and $k$ nearest nonimal neighbors \cite{SPADE}. 
To notably reduce the inference cost, PatchCore \cite{PatchCore} subsamples the memory bank. Moreover, Patchcore extracts locally aware patch features to achieve high performance. 
In this paper, the proposed POUTA is compared with PatchCOre, known for its efficiency, among the above methods.
The embedding-based methods generally enjoys higher efficiency than the reconstruction-based methods, but the anomaly locations are generally not as fine-grained as the latter.

\noindent \textbf{Reconstruction-based methods.} The reconstruction-based anomaly methods reconstruct a normal version of the input, typically through the autoencoder \cite{Memory2}. Since the anomaly is located by analyzing the pixel-level differences between the images before and after reconstruction, the anomaly prediction is always fine-grained.
Recently, there is a paradigm to analyze the reconstruction errors by a discriminative network \cite{DRAEM,RSTD,DeSTSeg,DSR,Lv}. Namely, the vanilla framework comprises a reconstructive network and a discriminative network in series. 
The input of the discriminative network contains only the input image and the reconstructed image. However, the procedure that compresses the representations to build a reconstructed image inevitably results in information loss, which may ultimately lead to inconsistencies between the reconstruction errors and the anomalies.
Lots of effort has been invested in enhancing the quality of the reconstructed image to ensure that the reconstruction errors provide more precise anomaly information. And most turns to incorporate an additional module into the reconstructive network, such as a memory bank \cite{Memory1,Memory2}, clustering \cite{AFEAN} or codebook \cite{DSR}.
Turning the reconstruction sub-task to the image inpainting task, where the input is gridded and then reconstructed in patch \cite{RIAD, Intra, SSPACB, SSMCTB}, also facilitates the construction of a high-quality reconstructed image. 
It also proposes to utilize the mathematical expectation map of the training dataset as the supplement of normal appearance \cite{DBPI}, aiming to assist the reconstructive network to build a high-quality normal image.
The above methods aim to build a high-quality reconstructed image, which follows the vanilla idea to locate the anomaly by analyzing the pixel-level reconstruction errors. It is costly but still can not deal with certain anomaly, such as the color-close-to-normal one. 
Aiming at improving both the accuracy and efficiency, it is necessary to propose a new paradigm. Our method opts to reuse the reconstructive representations, which provides more precise anomaly information, endowing our method to achieve better performance with less costs than the vanilla method.

\noindent \textbf{Few-shot anomaly detection.} Few-shot anomaly detection aims to indicate the anomaly with only a few nominal samples as the training support \cite{RegAD}. 
TDG \cite{TDG} incorporates a hierarchical generative module to model the distribution of multi-scale patches. To enhance the representations, TDG uses image transformations and optimizes discriminators to distinguish between real and fake patches, as well as between different transformations applied to the patches. The anomaly predictions are obtained by aggregating the patch-based votes of the correct transformations. Similarly, RegAD \cite{RegAD} applies the transformations as well, but on the features, learning the category-agnostic feature registration flexibly. Therefore, RegAD can detect anomaly in new categories with just a few normal images, without requiring fine-tuning \cite{RegAD}. 
MSMC \cite{MSMC} proposes to employs a global memory bank and an individual memory bank to record the features of the entire image and a single object respectively. The prior few-shot anomaly detection methods implement focused design for the few-shot context, whereas our method POUTA is initially designed for the typical anomaly detection. Despite this, POUTA still exhibits considerable performance on the few-shot anomaly detection, indicating strong learning ability towards the representations inherent in the training data.

\section{Method}
\begin{figure*}[h]
\begin{center}
    \includegraphics[width=\textwidth]{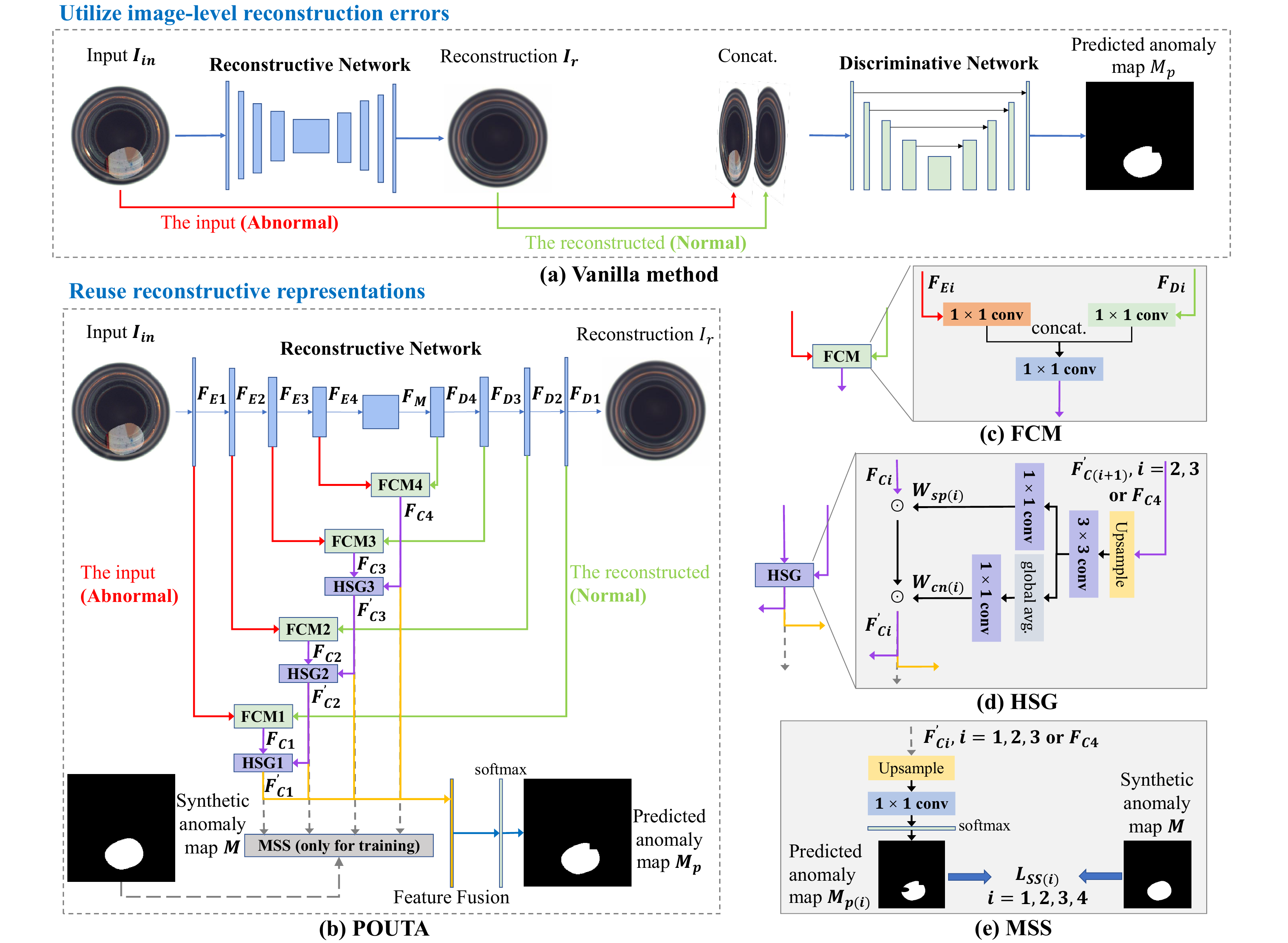}
\end{center}
   \caption{The vanilla method and the proposed POUTA. Vanilla method is composed of an autoencoder-like reconstructive network and a U-Net\cite{UNet}-like discriminative network, where the discriminative network obtains only the concatenation of the input and reconstructed image. While in POUTA, the discriminative network obtains the representations of the symmetric layers in the reconstructive network. The reconstructive representations contains more precise information about anomaly than the pixel-level reconstruction errors, which enables POUTA to identify anomaly more precisely.
   Each layer (except the first one) in the discriminative network involves a coarse(FCM module)-to-fine(HSG and MSS module) process.}
\label{fig:POUTA}
\end{figure*}

\noindent As depicted in Figure \ref{fig:POUTA}, instead of analyzing the pixel-level reconstruction errors in the vanilla method, 
POUTA predicts the anomaly by analyzing the discrepancies in the symmetric reconstructive representations. POUTA is a neat and effective framework which comprises two networks that we will describe in sequence: self-supervised learning with reconstructive network (\ref{sec:reconstructive_network}) and coarse-to-fine discriminative network (\ref{sec:discriminative_network}).

\subsection{Self-supervised learning with reconstructive network}
\label{sec:reconstructive_network}
Different from conventional methods, the reconstructed image does not need to repair the anomaly flawlessly, and it is sufficient to offer a tendency towards repairing it in POUTA. 
Therefore, no additional elements are added to produce a flawless reconstructed image, and the reconstructive network is requested to rectify the synthetic anomaly using contextual information, which has been proven to be an effective strategy to enhance the representational learning ability \cite{FCDD,Old,CutPaste}. 
Inspired by \cite{memseg}, the synthetic abnormal image $I_{in}$ is generated by seamlessly blending an augmented image $I_{aug}$ within a predetermined region $M$ of a training image $I_{ori}$. $I_{aug}$ is formed by merging a transformed version of the original image (e.g. translation, rotation or identity) with one obtained from an unrelated dataset \cite{DTD}. $M$ is specified by Perlin generator \cite{Perlin}. The input in Figure \ref{fig:POUTA} exemplifies the synthetic image.

There are three parts in reconstructive network: the encoder, the mapping layer, and the decoder. 
Given the input image, the encoder outputs hierarchy representations $F_{Ei}$ $(i=1,2,3,4)$. Then, the mapping layer maps the representation $F_{E4}$ to the normal latent space, and outputs as $F_{M}$. Subsequently, the decoder samples up $F_{M}$ and obtains the multi-scale normal representations $F_{Di}$ $(i=4,3,2,1)$. Finally, an image with normal appearance $I_r$ is reconstructed. 
$I_r$ is supposed to be the same with the original one $I_{ori}$. Therefore, the reconstructive loss $L_{rec}$ is calculated between $I_r$ and $I_{ori}$ by the mean square error (MSE) and structure similarity index measure \cite{SSIM} (SSIM):
\begin{equation}
    L_{rec} = L_{MSE}(I_{r},I_{ori}) + L_{SSIM}(I_{r},I_{ori})
\end{equation}

\subsection{The Coarse-to-fine discriminative network}
\label{sec:discriminative_network}
We find that the reconstructive representations contain a relatively comprehensive anomaly information, and some of the information may be lost during the compression of representations into images, resulting in the incomplete or even inaccurate anomaly information in the pixel-level reconstruction errors.
Previous methods focus on enhancing the quality of the reconstructed image by using sophisticated skills~\cite{DBPI,Memory1}, which inevitably requires extra space or computing resources. 
Instead, the reconstructive representations are employed as the discriminative representations directly in this paper. 
Our method endows the discriminative network with comprehensive anomaly information, and simultaneously improves detection efficiency by reusing the representations.

We propose a coarse-to-fine strategy to analyze the discrepancies between the symmetric reconstructive representations in POUTA. Three well-designed modules are 
proposed to detect and refine the anomaly location.  Specifically, the Feature Contrast Module (FCM) initially acquires the anomaly location, and then the High-level Semantic Guidance (HSG) and Multi-scale Semantic Supervision (MSS) refines the location.

\textbf{FCM.} 
FCM treats the encoder representations $F_{Ei}$ $(i=1,2,3,4)$ and the decoder representations $F_{Di}$ $(i=1,2,3,4)$ as the features of the input and reconstructed image respectively. By analyzing the discrepancies between them, the preliminary anomaly information exists in the contrastive feature maps $F_{Ci}$ $(i=1,2,3,4)$.
Specifically, as demonstrated in Figure \ref{fig:POUTA}, $F_{Ei}$ and $F_{Ci}$ $(i=1,2,3,4)$ go through their respective $1\times1$ convolutional layers. Next, a $1\times1$ convolutional layer is used to capture the discrepancies (distribution shift) between them by their channel-wise concatenation, and its output is denoted as $F_{Ci}$. $F_{Ci}$ contains the roughly precise information about the anomaly location.

\textbf{HSG.}
The anomaly information in $F_{Ci}$ is rough. HSG is designed to further refine it. And the refined version is denoted as $F^{\prime}_{Ci}$ $(i=1,2,3)$.
Generally speaking, the shallow representations (high-resolution) contains the information about anomaly details, but they also suffers interference from the noise. Although the deep representations (low-resolution) generally cannot provide detailed anomaly information, they reveal accurate anomaly semantics and are less affected by noise.
To refine the anomaly information contained in $F_{Ci}$ $(i=1,2,3)$, a natural idea is to filter the shallow $F_{Ci}$ with the deep one $F_{C(i+1)}$.

As illustrated in Figure \ref{fig:POUTA}, the last refined representation $F^{\prime}_{C(i+1)}$ $i=2,3$ or $F_{C4}$ are used to filter the current representation $F_{Ci}$ $(i=1,2,3)$. Specifically, the deep representation $F^{\prime}_{C(i+1)}$ $i=2,3$ (or $F_{C4}$) is firstly upsampled to the current resolution, and then passes through a $3\times3$ convolution. Next, it generates weight maps at spatial dimension $W_{sp(i)}$ and channel dimension $W_{cn(i)}$ as filters. To generate the spatial weight map $W_{sp(i)}$, a $1\times1$ convolution is added to adjust the channel number to $1$. As for the channel-wise weight map $W_{cn(i)}$, a global average pooling is used to adjust the spatial size to $1\times1$, and then $W_{cn(i)}$ is generated by a $1\times1$ convolution.
By multiplying $W_{sp(i)}$ and $W_{cn(i)}$ to $F_{C(i)}$, a refined contrastive feature map $F^{\prime}_{C(i)}$ is obtained:
\begin{equation}
    W_{sp(i)} = 
    \left\{
        \begin{aligned}
            & f_{sp(i)}(F^{\prime}_{C(i+1)}), & \quad i = 1,2 \\
            & f_{sp(i)}(F_{C4}), & i = 3
        \end{aligned}
        \right.
\end{equation}
\begin{equation}
    W_{cn(i)} = 
    \left\{
        \begin{aligned}
            & f_{cn(i)}(F^{\prime}_{C(i+1)}), & \quad i = 1,2 \\
            & f_{cn(i)}(F_{C4}), & i = 3
        \end{aligned}
        \right.
\end{equation}
\begin{equation}
    F^{\prime}_{Ci} = W_{sp(i)} \odot F_{Ci} \odot W_{cn(i)}, \quad
    i = 1,2,3
\end{equation}
where the $f_{sp(i)}$ represents the operations of upsample and convolution, and $f_{cn(i)}$ represents the operations of upsample, convolution and global average pooling. $F^{\prime}_{Ci}$ is involved to refine the next contrastive
feature map $F_{C(i-1)}$ $(i=2,3)$.

\textbf{MSS.}
POUTA establishes on the observation that the representations of the encoder and decoder in the reconstructive network refer to the features of the input and reconstructed image respectively. Under this observation, we further propose a multi-scale semantic supervision (MSS) module to refine the anomaly information by ensuring a clear discrepancy between the symmetric reconstructive representations.
MSS requires $F^{\prime}_{Ci}$ $(i=1,2,3)$ and $F_{C4}$ to be able to predict the abnormal regions $M_{p(i)}$ through their corresponding predictive head. As shown in Figure \ref{fig:POUTA}, the predictive head consists of an upsample layer, a $1\times1$ convolution for dimensional reduction and a softmax layer. Note that MSS module is only used during training.
To supervise the anomaly prediction of each layer, we add semantic supervision loss $L_{SS(i)}$ to each predictive head, which is represented as:  
\begin{equation}
    L_{SS(i)}\!=\!L_{focal}(M_{p(i)},M)\!+\!L_1(M_{p(i)},M), \; i=1,2,3,4
\end{equation}
where ${L_1}\left(  \cdot  \right)$ denotes the ${\ell _1}$-norm-based loss function, $L_{focal} \left(  \cdot  \right)$ is the focal loss \cite{FocalLoss},
$M$ is a binary map for ground truth, which indicates the abnormal regions.
MSS forces the intermediate representations to pay more attention to anomaly information. 
The total loss of MSS $L_{MSS}$ is the weighted sum of $L_{SS(i)}$ on each scale:
\begin{equation} \label{equ:LMSS}
    L_{MSS} = \sum_{i=1}^{4} \lambda_{i}L_{SS(i)}
\end{equation}

Finally, the channel-wise aggregation of $F^{\prime}_{C1}$, $F^{\prime}_{C2}$, $F^{\prime}_{C3}$ and $F_{C4}$ is required to predict a pixel-level anomaly heatmap $M_{p}$. Inspired by the vanilla method \cite{DRAEM}, the image-level anomaly score $S$ is the maximum of the $21\times 21$ average pooling result of $M_{p}$. The loss of the predicted heatmap $L_{pre}$ and the total loss of POUTA $L_{total}$ are represented as follows:
\begin{equation}
    L_{pre} = L_{focal}(M_{p},M) + L_1(M_{p},M)
\end{equation}
\begin{equation}
    L_{total} = L_{rec} + L_{pre} + L_{MSS}
\end{equation}

\section{Experiment}
\subsection{Experimental Setup}
\textbf{Dataset.}
POUTA is evaluated on MVTec AD \cite{MVTec}, VisA \cite{VisA}, and DAGM \cite{DAGM} dataset. MVTec AD is a benchmark dataset for visual anomaly detection, which contains 5,354 images in 15 image categories. VisA contains 10,821 images with 9,621 normal and 1,200 abnormal samples. Since 
abnormal images only exist in the testing set for MVTec AD and VisA, they are mainly used to train unsupervised anomaly detection methods. Different with the above datasets, DAGM has defect images in training set and is mainly used to train supervised anomaly detection methods.
It contains 16100 images in 10 categories. 
Note that POUTA doesn't use any defect images on all of the datasets, and it is trained on the defect-free images and the synthetic anomalies.

\textbf{Metric.}
In order to have a fair comparison with other methods, both the image-level classification performance (CLS) and pixel-level location performance (LOC) are exploited in this paper. The metric to measure CLS and LOC is the area under the receiver operating characteristic curve (AUROC). The average precision (AP) is also used to measure LOC. 
On MVTec AD dataset, both CLS and LOC are adopted to evaluate different methods. Since the annotations of DAGM is coarse, only the CLS is adopted to evaluate different methods. Similar with other state-of-the-art methods, CLS is adopted to evaluate different methods on VisA.

\textbf{Implementation Details.}
$\lambda_1$, $\lambda_2$, $\lambda_3$ and $\lambda_4$ in Equation~\ref{equ:LMSS} are empirically set to $0,4$, $0.3$, $0.2$ and $0.1$ respectively. 
POUTA is trained for 600 epochs with a batch size of 8. The learning rate of the Adam optimizer is set to 0.0002 and multiplied by 0.2 at epoch 480 and epoch 540. Images are resized to $224 \times 224$.

\subsection{Anomaly detection}
The proposed POUTA is compared with previous methods on MVTec AD, VisA and DAGM. The comparison results are shown in Table \ref{tab:MVTec} , \ref{tab:VisA} and \ref{tab:DAGM}.

\begin{table*}[h]
\begin{center}
\caption{Image-level classification (CLS) and pixel-level location (LOC) (in \%) of the vanilla method DR{\AE}M \cite{DRAEM}, several state-of-the-art reconstruction-based methods (including RSTD \cite{RSTD}, ReContrast \cite{ReContrast}, DeSTSeg \cite{DeSTSeg}, DBPI \cite{DBPI}), embedding-based method PatchCore \cite{PatchCore} and our method POUTA on MVTec AD dataset. The best result for each class is highlighted in bold.}
  \resizebox{\textwidth}{!}{
    \begin{tabular}{|l|cc|cc|cc|cc|cc|cc|cc|}
    \hline
    \multirow{2}{*}{class}
    & \multicolumn{2}{c|}{DR{\AE}M} & \multicolumn{2}{c|}{RSTD} & \multicolumn{2}{c|}{ReContrast} & \multicolumn{2}{c|}{DeSTSeg} & \multicolumn{2}{c|}{DBPI} & \multicolumn{2}{c|}{PatchCore} & \multicolumn{2}{c|}{POUTA} \\
    \cline{2-15}
    {} & {CLS} & {LOC} & {CLS} & {LOC} & {CLS} & {LOC} & {CLS} & {LOC} & {CLS} & {LOC} & {CLS} & {LOC} & {CLS} & {LOC} \\
    \hline
    carpet & 97.0 & 95.5/53.5    & 98.7 & \textbf{99.2}/-     & \textbf{99.8} & -/-     &  - & 96.1/72.8     & 99.7 & 98.7/{\textbf{80.6}}    & 98.5 & 99.0/63.9    & 99.1 & 98.0/77.2\\
    grid  & 99.9 & \textbf{99.7}/65.7     & 100.0 & 99.6/-    & 100.0 & -/-    & - & 99.1/61.5    & 99.8 & 99.6/69.1    & 98.2 & 98.5/31.5     & \textbf{100.0} & 99.2/\textbf{70.8}\\
    leather & 100.0 & 98.6/75.3     & 100.0 & 99.6/-    & 100.0 & -/-   & - & \textbf{99.7}/75.6    & 100.0 & 99.4/\textbf{76.0}    & 100.0 & 99.3/46.2     & \textbf{100.0} & 98.4/71.0\\
    tile  & 99.6 & 99.2/92.3     & 99.9 & 98.8/-     & 99.8 & -/-   & - & 98.0/90.0    & 100.0 & 99.5/96.9    & 99.1 & 95.8/55.3    & \textbf{100.0} & \textbf{99.7}/\textbf{97.9}\\
    wood  & 99.1 & 96.4/77.7    & 99.3 & 98.1/-    & 99.0 & -/-     & - & 97.7/81.9     & 99.7 & 96.9/78.5     & 99.2 & 95.1/50.1     & \textbf{100.0} & \textbf{98.2}/\textbf{85.4}\\ 
    bottle & 99.2 & 99.1/86.5    & 100.0 & 99.3/-    & 100.0 & -/-     & - & 99.2/90.3    & 99.9 & 98.9/88.0    & 100.0 & 98.5/77.5     & \textbf{100.0} & \textbf{99.3}/\textbf{91.6}\\
    cable & 91.8 & 94.7/52.4     & 99.6 & 98.3/-     & \textbf{99.8} & -/-     & - & 97.3/60.4     & 95.7 & 96.7/66.8    & 99.3 & 98.1/66.5     & 97.8 & \textbf{98.4}/\textbf{75.9}\\
    capsule & 98.5 & 94.3/49.4    & 93.0 & 98.5/-     & 97.7 & -/-    & - & 99.1/\textbf{56.3}     & 97.1 & 98.6/48.2    & 96.7 & 99.0/44.5    & \textbf{99.9} & \textbf{99.1}/54.7\\
    hazelnut & 100.0 & \textbf{99.7}/\textbf{92.9}     & 99.8 & 99.5/-     & 100.0 & -/-    & - & 99.6/88.4     & 100.0 & 99.5/89.4    & 100.0 & 98.6/52.3     & \textbf{100.0} & 99.6/88.0\\
    metal nut & 98.7 & \textbf{99.5}/96.3     & \textbf{100.0} & 98.9/-     & \textbf{100.0} & -/-     & - & 98.6/93.5     & 98.9 & 98.0/{92.2}    & 99.7 & 98.3/87.0     & 99.7 & 99.3/\textbf{97.1}\\
    pill  & \textbf{98.9} & 97.6/48.5     & 98.1 & 98.7/-     & 98.6 & -/-     & - & 98.7/83.1     & 96.6 & \textbf{99.4}/88.2    & 98.0 & 97.6/78.6     & 98.0 & 99.3/\textbf{88.8}\\
    screw & 93.9 & 97.6/58.2     & 96.8 & \textbf{99.3}/-     & 98.0 & -/-     & - & 98.5/58.7     & 98.4 & 99.3/\textbf{60.7}    & 97.2 & \textbf{99.4}/37.2     & \textbf{98.5} & 97.7/58.1 \\
    toothbrush & 100.0 & 98.1/44.7     & 97.9 & 99.3/-     & 100.0 & -/-     & - & 99.3/75.2     & 100.0 & 99.3/78.1    & 99.4 & 98.5/37.4     & \textbf{100.0} & \textbf{99.5}/\textbf{82.8}\\
    transistor & 93.1 & 90.9/50.7     & 98.3 & 90.7/-     & 99.7 & -/-     & - & 89.1/64.8     & 99.1 & 95.3/72.5    & 100.0 & 95.1/65.0     & \textbf{100.0} & \textbf{97.9}/\textbf{79.5}\\
    zipper & 100.0 & 98.8/81.5     & 99.3 & \textbf{99.2}/-     & 99.5 & -/-     & - & 99.1/\textbf{85.2}     & 98.8  & 98.6/74.1     & 99.5 & 98.8/62.3     & \textbf{100.0} & 98.9/83.5\\
    \hline
    avg   & 98.0 & 97.3/68.4     & 98.7 & 98.5/-     & 99.5 & 98.4/-     & - & 97.9/75.8     &98.9  & 98.5/77.3     & 99.0 & 98.0/57.0     & \textbf{99.5} & \textbf{98.8}/\textbf{80.2}\\
    \hline
    \end{tabular}%
    }
\end{center}
  \label{tab:MVTec}%
\end{table*}%

\begin{figure*}[h]
\begin{center}
    \includegraphics[width=\textwidth]{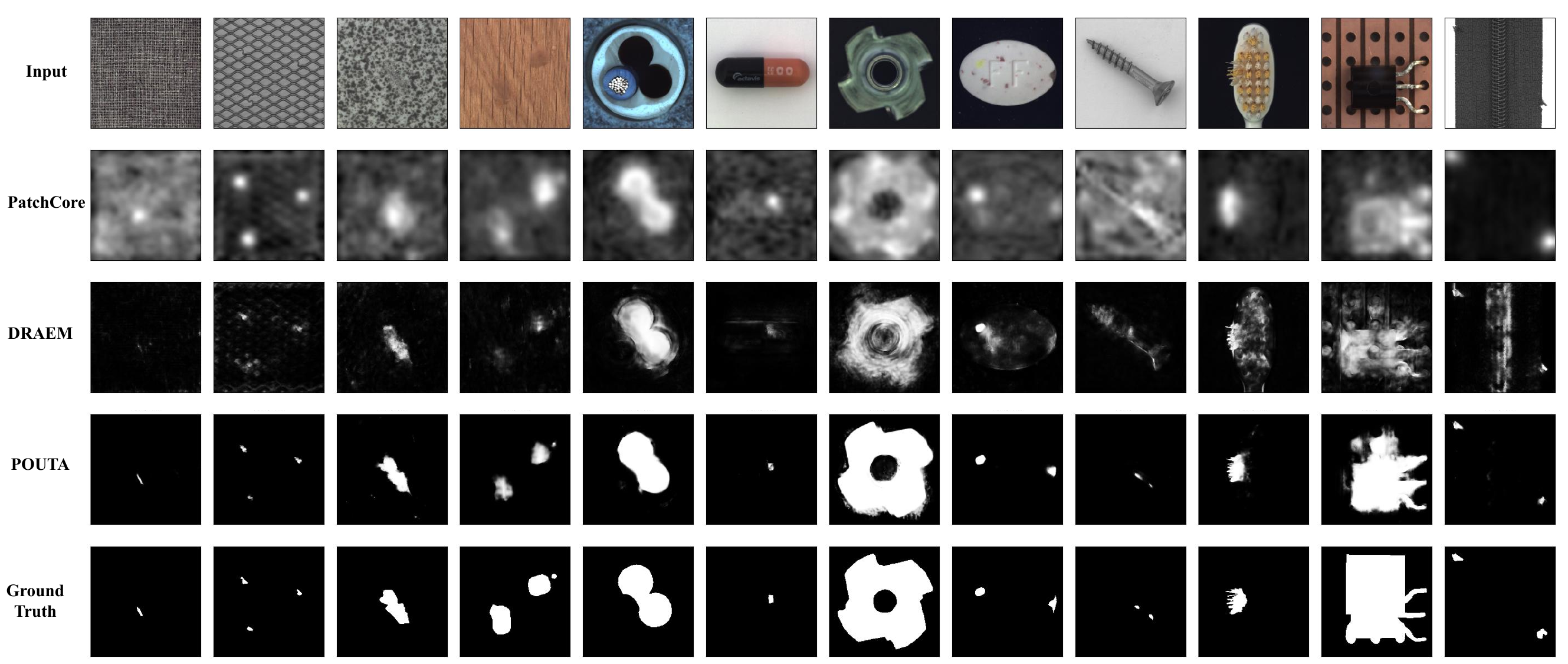}
\end{center}
   \caption{The qualitative results of DRAEM \cite{DRAEM}, PatchCore \cite{PatchCore}, and POUTA on MVTec AD \cite{MVTec}.}
\label{fig:AD_MVTec}
\end{figure*}

\begin{table*}[h]
\caption{Image-level classification (CLS) (in \%) of several state-of-the-art reconstruction-based methods (including EdgRec \cite{EdgRec}, RD(Reverse Distillation) \cite{reverse}, ReContrast \cite{ReContrast}), embedding-based methods (PatchCore \cite{PatchCore}), and our POUTA on VisA dataset (AUROC). The best result is highlighted in bold.}
\begin{center}
  \resizebox{\textwidth}{!}{
    \begin{tabular}{c|cccccccccccc|c}
    \hline
    {} & {PCB1} & {PCB2} & {PCB3} & {PCB4} & {Capsules} & {Candle} & {Macaroni1} & {Macaroni2} & {Cashew} & {Chewing gum} & {Fryum} & {Pipe fryum} & {avg.} \\
    \hline
    {EdgRec} & - & - & - & - & - & - & - & - & - & - & - & - & {94.2} \\
    {RD} & - & - & - & - & - & - & - & - & - & - & - & - & {96.0} \\
    {ReContrast} & - & - & - & - & - & - & - & - & - & - & - & - & {97.5} \\
    {PatchCore} & \textbf{98.5} & {97.1} & {98.4} & \textbf{99.8} & {76.2} & \textbf{98.9} & \textbf{97.2} & 78.6 & \textbf{97.9} & 98.5 & 94.5 & 99.7 & {94.6} \\
    {POUTA} & {97.0} & \textbf{99.6} & \textbf{98.4} & {99.1} & \textbf{97.2} & {96.9} & {95.7} & \textbf{95.7} & {96.4} & \textbf{99.7} & \textbf{98.8} & \textbf{99.7} & \textbf{97.9} \\
    \hline
    \end{tabular}%
    }
\end{center}
  \label{tab:VisA}%
\end{table*}%

\begin{figure*}[h]
\begin{center}
    \includegraphics[width=\textwidth]{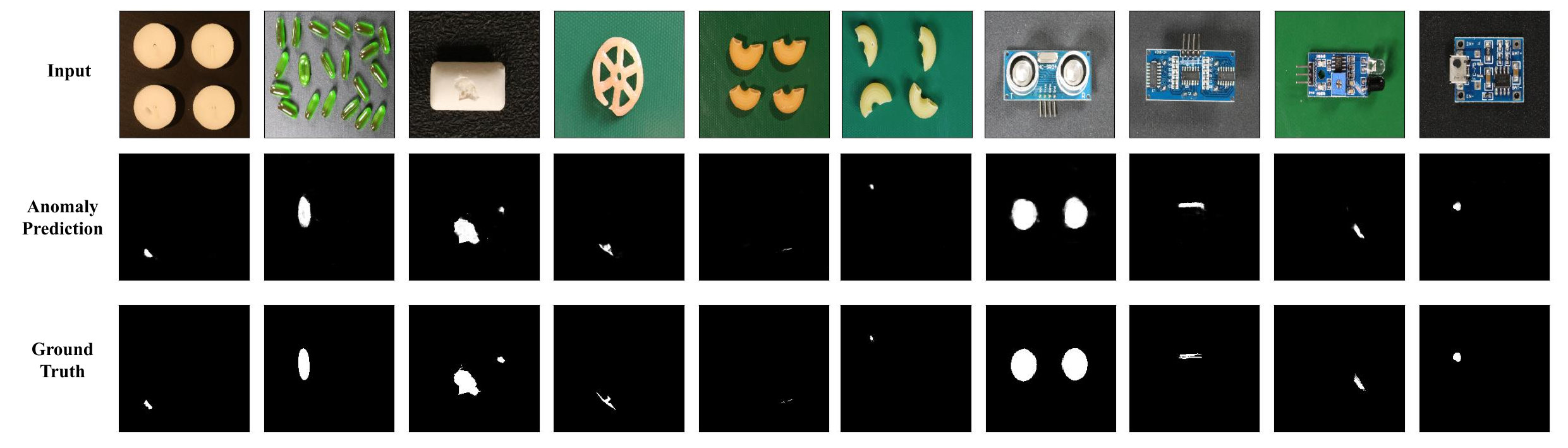}
\end{center}
   \caption{The qualitative results of POUTA on VisA dataset\cite{VisA}.}
\label{fig:AD_VisA}
\end{figure*}

\begin{table*}[h]
\caption{Image-level classification (CLS) (in \%) of several state-of-the-art reconstruction-based methods (including DBPI \cite{DBPI} and DR{\AE}M \cite{DRAEM}), embedding-based method (PatchCore \cite{PatchCore}), supervised methods (including CCNN \cite{CCNN} and MS \cite{MS}), and our POUTA on DAGM dataset (AUROC). The best result is highlighted in bold.}
\begin{center}
    \begin{tabular}{c|c|cccccccccc|c}
    \hline
    & {} & {Class1} & {Class2} & {Class3} & {Class4} & {Class5} & {Class6} & {Class7} & {Class8} & {Class9} & {Class10} & {avg.} \\
    \hline
    \multirow{4}{*}{\rotatebox{90}{unsup.}}
    & {DBPI} & - & - & - & - & - & - & - & - & - & - & {99.0} \\
    & {DR{\AE}M} & - & - & - & - & - & - & - & - & - & - & {96.0} \\
    & {PatchCore} & 87.3 & 100.0 & 93.5 & \textbf{100.0} & 91.4 & 97.9 & 99.8 & 61.4 & 95.4 & 100.0 & {92.7} \\
    & {POUTA} & \textbf{100.0} & \textbf{100.0} & \textbf{100.0} & {99.9} & \textbf{99.9} & \textbf{100.0} & \textbf{100.0} & \textbf{100.0} & \textbf{97.4} & \textbf{100.0} & \textbf{99.7} \\
    \hline
    \multirow{2}{*}{\rotatebox{90}{sup.}}
    & {CCNN} & {99.0} & {100.0} & {98.0} & {99.0} & {100.0} & {100.0} & {100.0} & {100.0} & {100.0} & {100.0} & {99.6} \\
    & {MS} & - & - & - & - & - & - & - & - & - & - & \textbf{100.0} \\
    \hline
    \end{tabular}%
\end{center}
  \label{tab:DAGM}%
\end{table*}%

\textbf{MVTec AD.}
We report the quantitative classification (CLS) and location (LOC) results of the prior methods and POUTA on MVTec AD dataset in Table \ref{tab:MVTec}. PatchCore in this section is conducted using its official code with 1\% training data as coreset remaining.
As for the CLS results, POUTA shows highest performance on most cases. 
Especially on capsule, POUTA outperforms others by 1.4\%.
We also notice that the difference between POUTA and the best result is 2\% on cable, which may due to changes in cable's orientation, as well as the background. They prevent POUTA from modeling the normal appearance of cable, causing to  judge these changes as anomalies.
For the average performance of CLS, POUTA attains a comparable  performance with SOTA results.
And for the LOC results, POUTA also demonstrates satisfactory performance in the majority  cases, with little performance difference in the remaining cases. For the AUROC metric, POUTA surpasses the SOTA by 2.6\% on transistor, which is because of the more complete anomaly location predicted by POUTA, especially when the anomaly is large. The performance improvement is more clear on AP metric. On wood, cable, and transistor, POUTA outperforms the SOTA by 6.9\%, 9.1\%, and 7.0\%.
On the average LOC results, POUTA is 0.3\% and 2.9\% higher than SOTA measured by pixel-level AUROC and AP metric respectively. A qualitative comparison with the vanilla method (DRAEM \cite{DRAEM}) and embedding-based method (PatchCore \cite{PatchCore}) is presented in Figure \ref{fig:AD_MVTec}, indicating that POUTA is able to precisely predict the anomaly with more details.

\textbf{VisA.}
POUTA is further compared with other methods on VisA datasets. 
As shown in Table \ref{tab:VisA}, POUTA outperforms the prior methods on most cases. Especially on capsules and macaroni2, POUTA shows 21.0\% and 17.1\% improvement on the image-level AUROC metric. Both of these are the categories where the position of the object changes drastically per image. This indicates that POUTA is more flexible than others to adapt to these position changes.
And as for the average classification performance, POUTA is 0.4\% higher than the SOTA.
Sample segmentations in Figure \ref{fig:AD_VisA} offer a qualitative anomaly localization of POUTA, showing that POUTA can locate the anomaly accurately.

\textbf{DAGM.}
In addition, the quantitative evaluations on DAGM dataset are presented in Table \ref{tab:DAGM}. POUTA outperforms all the unsupervised methods 
on all the categories except for Class4.
It is worth mentioning that on Class8, POUTA is 38.6\% higher than PatchCore, which measures anomaly by calculating the distances between the test sample and the coreset's training samples. However, the texture of Class8 varies, which makes it to be impossible to include all possible textures in the training data, resulting in a non-ideal detection performance. While POUTA learns the distribution of the texture in the reconstructive network, which results in minimal texture changes before and after reconstruction, leading to more accurate detection results.
The average performance is also higher than the prior unsupervised methods.
Besides, POUTA achieves a comparable performance with the best supervised methods. 
While POUTA requires no real anomaly for training, which enables POUTA to be more flexible to the situation where the real anomaly is hard to collect.

\subsection{Few-Shot Anomaly Detection}
\begin{table}[h]
\begin{center}
\caption{Image-level few-shot classification (CLS) (AUROC in \%) of the prior few-shot anomaly detection methods (including TDG \cite{TDG}, DifferNet \cite{DifferNet}, RegAD \cite{RegAD}, and POUTA on MVTec AD dataset. The best result is highlighted in bold.}
    \begin{tabular}{lccc}
    \hline
    {} & {2-shot} & {4-shot} & {8-shot} \\
    \hline
    {TDG} & 71.2 & 72.7 & 75.2 \\
    {DifferNet} & 80.5 & 80.8 & 82.9 \\
    {RegAD} & 81.5 & 84.9 & 87.4 \\
    {POUTA} & \textbf{86.0} & \textbf{88.6} & \textbf{92.2} \\
    \hline
    \end{tabular}%
\end{center}
  \label{tab:FSAD_MVTec}%
\end{table}%

\begin{figure*}[h]
\begin{center}
    \includegraphics[width=\textwidth]{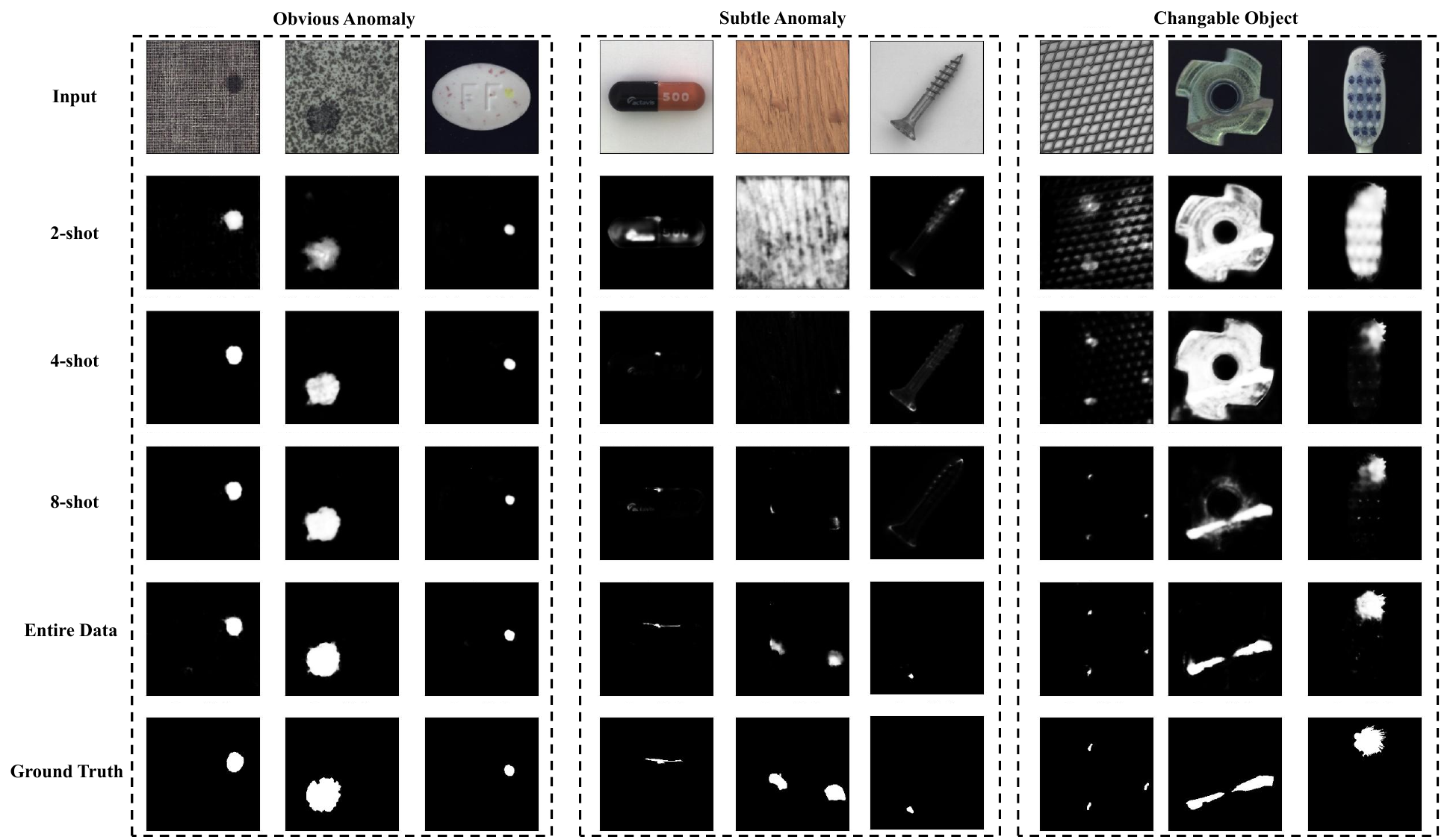}
\end{center}
   \caption{The qualitative few-shot and regular anomaly detection results of POUTA on MVTec AD\cite{MVTec}.}
\label{fig:FSAD}
\end{figure*}

\begin{figure}[h]
\begin{center}
    \includegraphics[width=\linewidth]{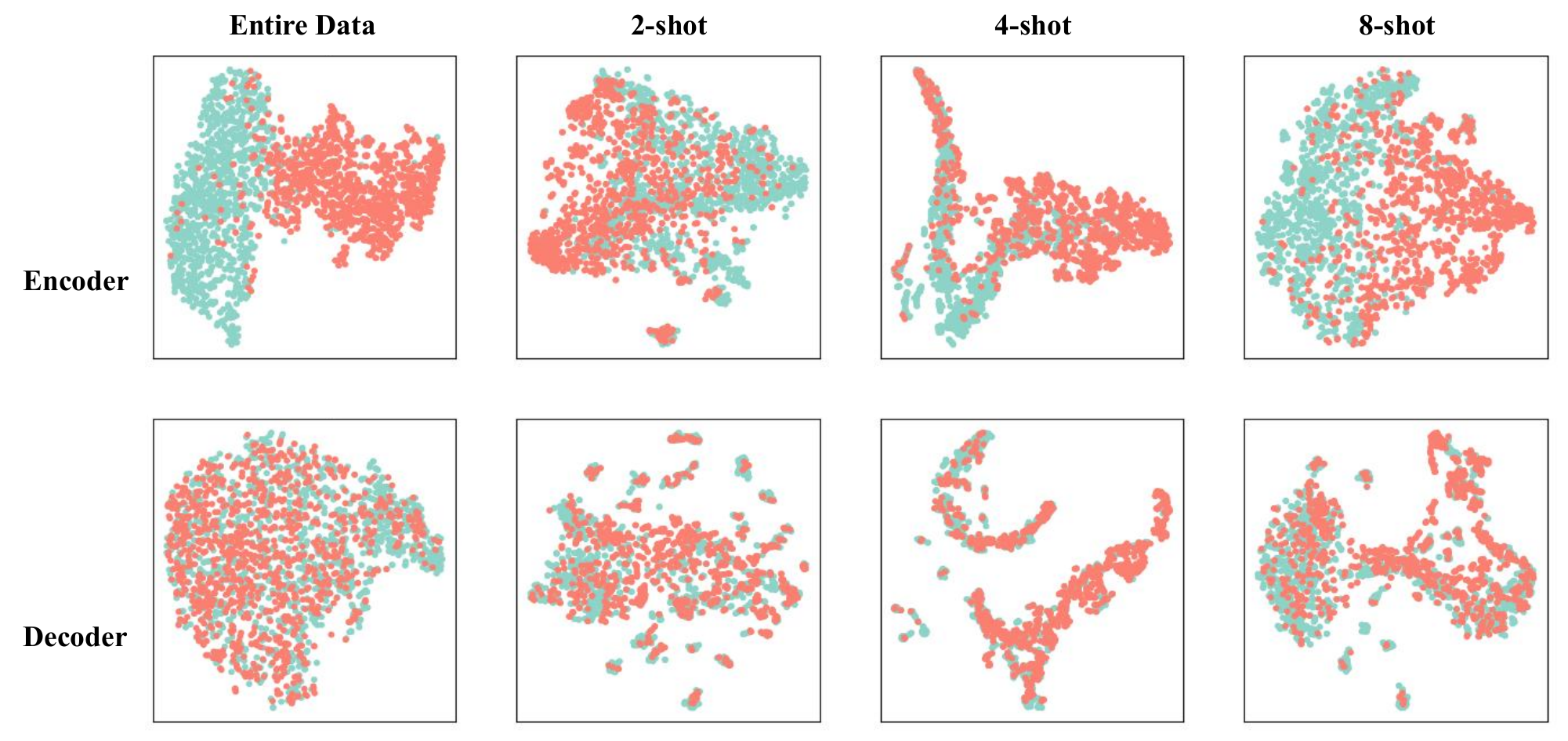}
\end{center}
   \caption{The visualization of the encoder and decoder representations of POUTA through t-SNE \cite{tSNE}.}
\label{fig:FSAD_tSNE}
\end{figure}

Few-shot anomaly detection becomes a hot topic recently since it is suitable for practical applications 
where the normal samples are hard to collect as well.  
Experiments are also done to analyze the effectiveness of our method on this task.
We train POUTA only with the support set ($k$ training images, $k=2,4,8$) for each category. Table \ref{tab:FSAD_MVTec} demonstrates the quantitative results of TDG \cite{TDG}, DifferNet \cite{DifferNet}, RegAD \cite{RegAD} and POUTA on MVTec AD. TDG and TegAD are designed for the few-shot anomaly detection methods, and DifferNet is an embedding-based anomaly detection methods with the normalizing flow. 

Table \ref{tab:FSAD_MVTec} indicates that POUTA achieves better performance than the prior few-shot anomaly detection methods, even without any special design. This is a surprising result, showing that POUTA has a strong ability to learn representations inherent in the training data.
In Figure \ref{fig:FSAD}, we demonstrate both the few-shot and regular anomaly detection of POUTA. The subtle anomaly tends to be close to the edge of the normal data distribution, requiring more training data to construct an accurate distribution. Similarly, it is difficult to develop an accurate distribution for a changeable object with few-shot training data. 
Thus even the results of POUTA are not perfect in the above situation.
While it can also be seen that, even with 2-shot, POUTA is still able to locate the obvious anomaly, indicating that POUTA is able to learn a near accurate distribution of the normal data.

The visualization in Figure \ref{fig:FSAD_tSNE} shows that with a few more training images (from 2-shot to 8-shot), the distribution established by POUTA is greatly improved, approaching to the distribution of entire data training. This confirms the strong ability to learn the inherent distribution of normal data.

\subsection{Ablation Study}
\label{sec:ablation}
To investigate the effect of each component in POUTA, five groups of ablation experiments are conducted on MVTec AD. All the experiments share the same implementation details.

\begin{table}[h]
    \caption{\quad Ablation study of POUTA on MVTec AD. The best result is highlighted in bold.}
    \centering
    \begin{tabular}{cccccc}
    \hline
    \multirow{2}{*}{Method} & \multicolumn{3}{c}{Module} & \multicolumn{2}{c}{average} \\
    \cline{2-4}
    {} & {FCM(Reuse)} & {HSG} & {MSS} & {CLS} & {LOC} \\
    \hline
    {Vanilla} & {} & {} & {} & 98.4  & 98.1/75.4 \\
    {POUTA-base} & {\Checkmark} & {} & {} & 98.8  & 98.1/76.7 \\
    {POUTA-base+HSG} & {\Checkmark} & \Checkmark & {} & 98.9  & 98.4/76.8 \\
    {POUTA-base+MSS} & {\Checkmark} & {} & {\Checkmark} & 98.8  & 98.3/77.2\\
    {POUTA} & {\Checkmark} & {\Checkmark} & \Checkmark & \textbf{99.5}  & \textbf{98.8}/\textbf{80.2} \\
    \hline
    \end{tabular}%
    \label{tab:Ablation}
\end{table}%

\begin{table}[h]
    \caption{\quad The calculation cost (FLOPs), parameters and average inference time per image of PatchCore, vanilla method, and POUTA. The best result is highlighted in bold.}
    \centering
    \begin{tabular}{cccc}
    \hline
    {}  & POUTA & Vanilla & PatchCore\\
    \hline
    {Calculation} & \textbf{35.37G} & {39.38G} & {-}\\
    {Parameters} & \textbf{15.02M} & {28.68M} & {-}\\
    {Inference time} & \textbf{32ms} & {52ms} & {38ms}\\
    \hline
    \end{tabular}%
    \label{tab:time}
\end{table}%

\begin{figure*}[h]
\begin{center}
    \includegraphics[width=\textwidth]{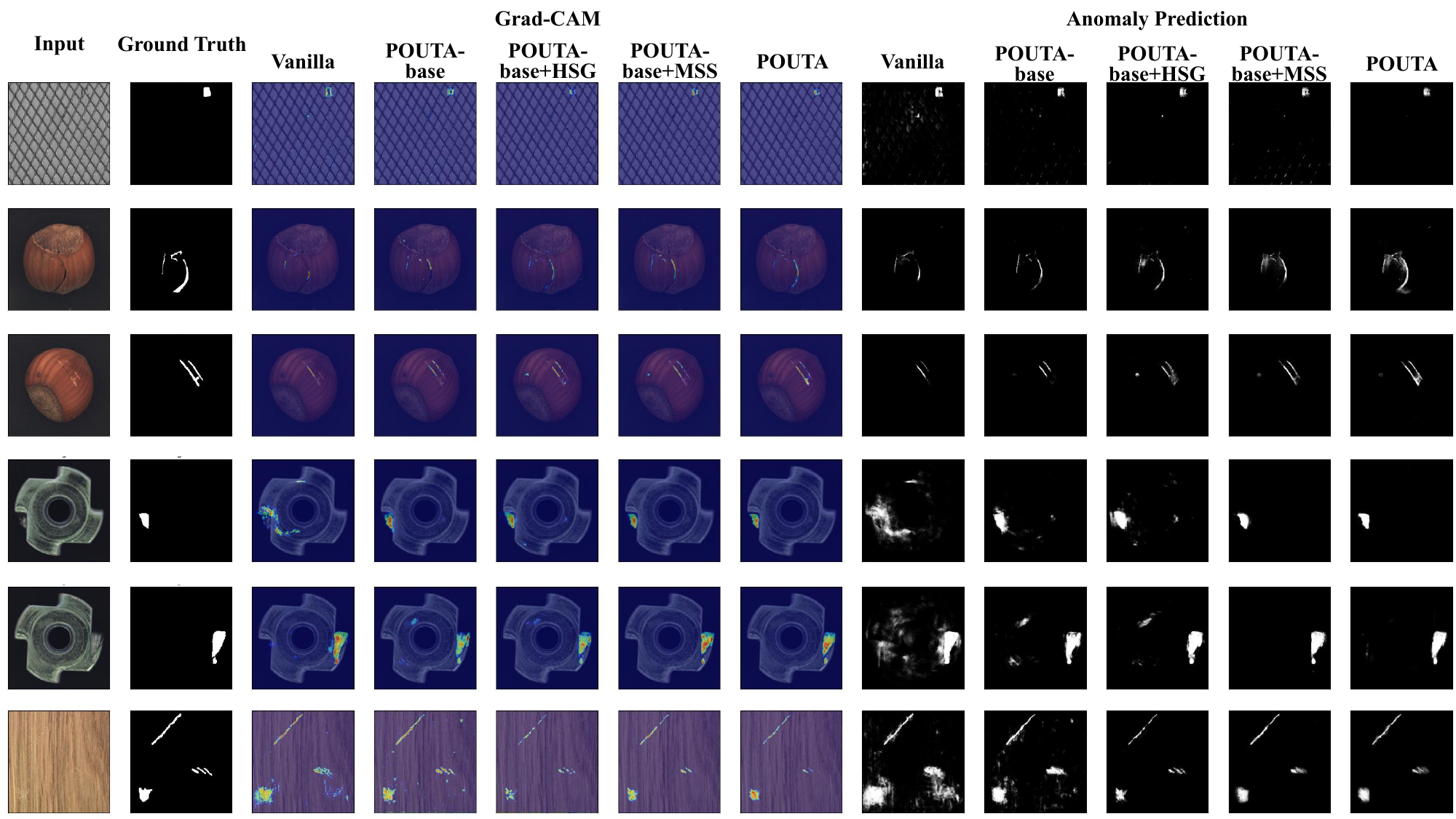}
\end{center}
   \caption{The qualitative examples and visualizations (proivded by Grad-CAM \cite{GradCAM}) of vanilla method, POUTA-base and POUTA. POUTA obtains a more comprehensive set of anomaly information to locate the anomaly more precisely.}
\label{fig:Ablation}
\end{figure*}

\begin{figure}[h]
\begin{center}
    \includegraphics[width=0.7\linewidth]{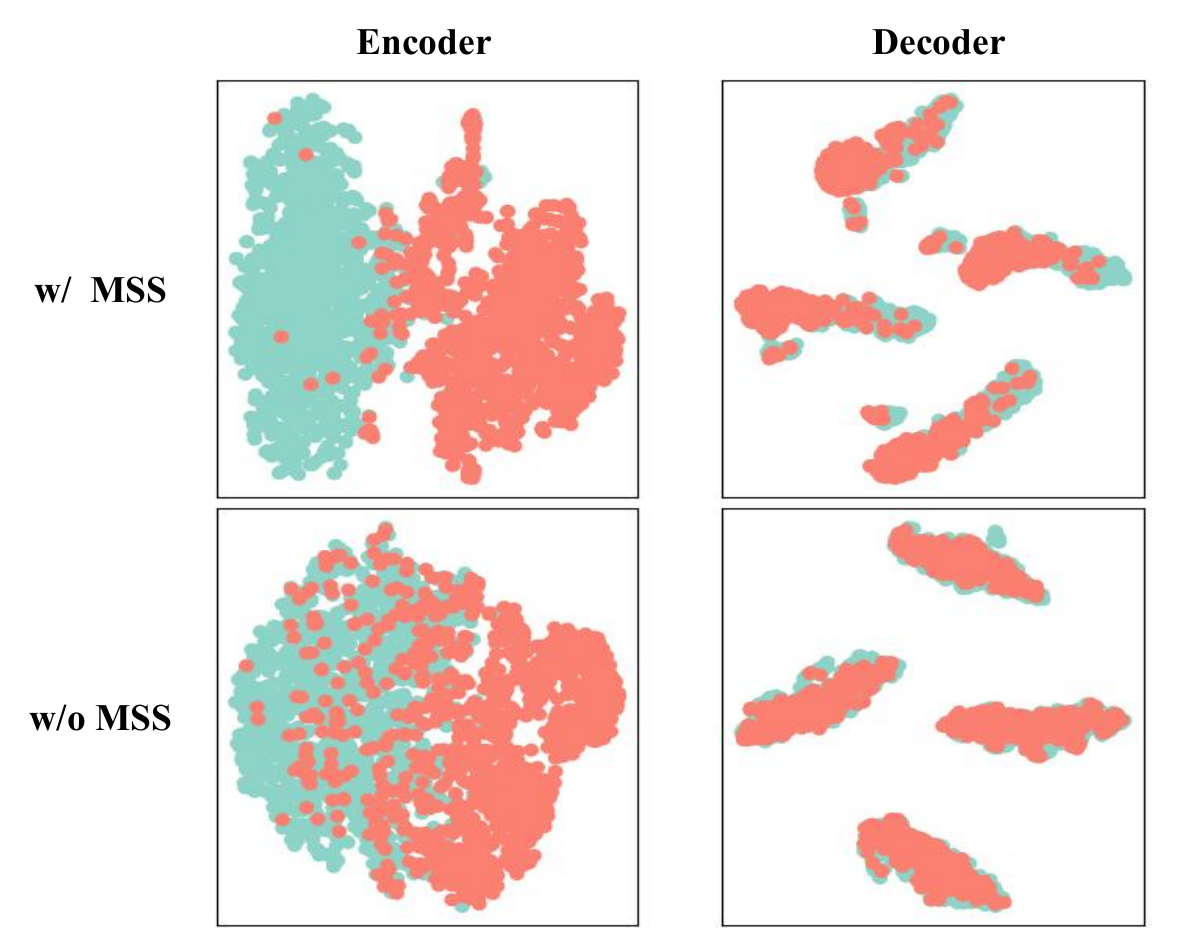}
\end{center}
   \caption{We take t-SNE \cite{tSNE} to visualize the representation distribution before and after mapping layer (the last encoder representations and the first decoder representations) in reconstructive network under with or without MSS module.}
\label{fig:Ablation_tSNE}
\end{figure}

\noindent \textbf{Quantitative Results.}
The quantitative results are shown in Table \ref{tab:Ablation}. 
\textit{Vanilla} refers to the vanilla method as shown in Figure \ref{fig:POUTA} (a), which utilizes only the pixel-level reconstruction errors to analyze the anomaly.
\textit{POUTA-base} is POUTA without HSG and MSS modules, that is, there is no refined step in the discriminative network and the coarse information from reconstructive network is directly used to locate the anomaly.
\textit{POUTA-base+HSG} and \textit{POUTA-base+MSS} add only HSG or MSS module to \textit{POUTA-base} respectively.
And 
\textit{POUTA} adds both HSG and MSS modules to \textit{POUTA-base}.

\begin{enumerate}
    \item \textbf{Reconstruction errors vs representation reusing:}
    Experiments \textit{Vanilla} and \textit{POUTA-base} in Table \ref{tab:Ablation} are conducted to verify
    the effectiveness of the representation reusing.
    The results shows that by reusing the reconstructive representations instead of the pixel-level reconstruction errors, the performance improves. This demonstrates that reusing representations from the reconstructive network provides more accurate anomaly information than the conventional reconstruction errors. 
    \item \textbf{HSG module:}
    To investigate the effectiveness of HSG, experiments \textit{POUTA-base} and \textit{POUTA-base+HSG} are conducted. \textit{POUTA-base+HSG} adds only HSG module to \textit{POUTA-base}, and the results show improvement in terms of the detection performance in Table \ref{tab:Ablation}. This shows that HSG module can be used to refine anomaly information.
    However, we also find that adding HSG only has marginal performance improvement. 
    Due to the lack of MSS, the representations of the encoder and the decoder cannot represent the features of the original and reconstructed image fully. 
    The semantic differences between the last few encoder representations and the first few decoder representations might not be obvious enough, resulting in unclear information about the abnormal regions at the high-level semantics, which affects the ability to guide the low-level representations.
    \item \textbf{MSS module:}
    \textit{POUTA-base+MSS} adds MSS module to \textit{POUTA-base}. As shown in Table \ref{tab:Ablation}, the detection performance increases by adding MSS module. 
    MSS requires each layer of discriminative network to predict the anomaly precisely, which enhances the ability to distinguish the normal and abnormal regions. While without HSG, each layer predicts the anomaly based on the representations of the current scale, which might be not robust and accurate enough since it has not been calibrated by the high-level semantic information. Therefore, adding only MSS module provides limited improvement.
    \item \textbf{The coarse-to-fine process:}
    \textit{POUTA} adds both HSG and MSS modules to \textit{POUTA-base}. There is a significant performance increase as shown in Table \ref{tab:Ablation}. \textit{POUTA} also shows a noticeable improvement when compared with \textit{POUTA-base+HSG} and \textit{POUTA-base+MSS}. The above results demonstrate that the collaboration between HSG and MSS plays an important role in refining the semantic information about the anomaly, allowing a better performance than adding only one or neither module.
    \item \textbf{Parameters, computation and inference time.}
    We calculate the average inference time per image on the embedding-based method \textit{PatchCore \cite{PatchCore}}, the reconstruction-based method \textit{Vanilla}, and the proposed method POUTA. Also, the computational cost (FLOPs and parameters of each model) are calculated. The results are shown in Table \ref{tab:time}. The parameters of POUTA are obviously less than those of \textit{Vanilla}, while the computation is slightly less than that of \textit{Vanilla}. This is because POUTA utilizes the existing information more extensively and refines it. It can be seen from the inference time that POUTA is significantly more efficient than \textit{Vanilla}, and slightly more efficient than \textit{PatchCore}.
\end{enumerate}

\noindent \textbf{Qualitative Results.}
The qualitative results are exhibited in Figure \ref{fig:Ablation}. Also, we take Grad-CAM \cite{gradcam} to visualize the semantic differences captured by the five groups of models in Figure \ref{fig:Ablation}.

\begin{enumerate}
    \item \textbf{Reconstruction errors vs representation reusing:}
    Figure \ref{fig:Ablation} demonstrates that the reconstructive representations enable the model to capture more accurate anomaly semantics, resulting in \textit{POUTA-base} to locate the anomaly more completely and precisely than \textit{Vanilla}.
    \item \textbf{HSG module:}
    As shown in Figure \ref{fig:Ablation}, the visualization of Grad-CAM shows that the addition of HSG enables the model to capture more accurate semantic differences, leading to predicting anomalies more precisely.
    This indicates that by introducing the high-level semantic information to the low-level feature through HSG module, it does benefit to calibrate the semantics. 
    But the gains are limited, since without MSS, the high-level semantic difference is not clear enough to describe the anomaly, which weakens the ability to guide the low-level representations.
    \item \textbf{MSS module:}
    As exemplified in Figure \ref{fig:Ablation}, \textit{POUTA-base+MSS} alleviates the misjudgment problem in \textit{POUTA-base} through MSS module.
    We take t-SNE \cite{tSNE} to visualize the representation distribution before and after mapping layer (the last encoder representations and the first decoder representations) in the reconstructive network with or without the MSS module, as shown in Figure \ref{fig:Ablation_tSNE}. It can be seen that, the main differences exist in the distributions of the encoder, where the encoder representations are significantly more separable with MSS, leading to a more obvious distribution shift between the encoder and decoder.
    The above phenomena verifies that MSS effectively improves the ability to distinguish the normal and abnormal information.
    However without HSG, the anomaly predictions of \textit{POUTA-base+MSS} are still not ideal enough in some cases as shown in Figure \ref{fig:Ablation}, indicating that even if MSS module is added, HSG module is still necessary to calibrate the semantics.
    \item \textbf{The coarse-to-fine process:}
    As presented in Figure \ref{fig:Ablation}, the anomaly predictions are significantly more precisely after adding HSG and MSS modules simultaneously. 
    With HSG, the semantic information of each layer in the discriminative network is more accurate, since it has been calibrated by the high-level features, further enabling MSS to achieve better supervision effect. And MSS enables the distribution discrepancies between encoder and decoder to be more obvious and makes the high-level semantic anomaly information to be more clear by the supervision loss. The collaboration between HSG and MSS refines the semantic information about the anomaly, allowing a significantly better performance than adding only one or neither module.
\end{enumerate}

\section{Conclusion}
In this paper, we verifies that reusing the reconstructive representations is more reliable than the pixel-level reconstruction errors for reconstruction-based anomaly detection. Then a new approach named POUTA is proposed locating the anomaly by analyzing the distribution shift between the reconstructive encoder and decoder,
which enables to predict the anomaly more precisely with lower cost than the vanilla method. 
Moreover, POUTA further refines the anomaly information by HSG and MSS modules in each discriminative layer, which also enhances the separability between the symmetric reconstructive representations.
Finally, POUTA outperforms the prior methods on MVTec AD, VisA and DAGM dataset, and achieves a comparable results with the best-performing supervised methods on DAGM dataset. 
Surprisingly, POUTA also achieves better performance than the state-of-the-art few-shot anomaly detection methods without any special design, showing that POUTA has strong ability to learn representations inherent in the training data.



\end{document}